\documentclass{article}
\usepackage[hyphens]{url}
\usepackage{hyperref}

\usepackage[preprint]{icml2026}

\usepackage{microtype}
\usepackage{graphicx}
\usepackage{booktabs}
\usepackage{subcaption}
\usepackage{multirow}
\usepackage{amsmath,amssymb,amsfonts}
\usepackage{bm}

\icmltitlerunning{The Value of Thinking: Learning When to Reason in Hybrid Decision Models}
\begin{document}

% This version names no corresponding author, so the footnote drops the
% sentence that would name one. `printAffiliationsAndNotice` is reproduced from
% icml2026.sty without its correspondence branch, which also removes the
% AUTHORERR the style prints when the macro is absent. The style file itself is
% left as the organisers published it.
\makeatletter
\renewcommand{\printAffiliationsAndNotice}[1]{\global\icml@noticeprintedtrue%
  \stepcounter{@affiliationcounter}%
  {\let\thefootnote\relax\footnotetext{\hspace*{-\footnotesep}\ificmlshowauthors #1\fi%
      \forloop{@affilnum}{1}{\value{@affilnum} < \value{@affiliationcounter}}{
        \textsuperscript{\arabic{@affilnum}}\ifcsname @affilname\the@affilnum\endcsname%
          \csname @affilname\the@affilnum\endcsname%
        \else
          {\bf AUTHORERR: Missing \textbackslash{}icmlaffiliation.}
        \fi
      }.%
      \ \\
      \Notice@String
    }
  }
}
\makeatother
\twocolumn[
\icmltitle{The Value of Thinking:\\Learning When to Reason in Hybrid Decision Models}

\begin{icmlauthorlist}
\icmlauthor{Zhengxu Yu}{noah}
\icmlauthor{Ke Xu}{liv}
\icmlauthor{Meng Fang}{liv}
\icmlauthor{Haitham Ammar}{noah}
\icmlauthor{Weilin Luo}{noah}
\icmlauthor{Jun Wang}{ucl}
\end{icmlauthorlist}

\icmlaffiliation{noah}{Huawei Noah's Ark Lab}
\icmlaffiliation{liv}{University of Liverpool}
\icmlaffiliation{ucl}{University College London}

\icmlkeywords{Machine Learning, ICML}

\vskip 0.3in
]

\printAffiliationsAndNotice{}

\begin{abstract}
  Existing adaptive-reasoning methods shorten the explicit reasoning a model writes, or
  skip it by switching modes; either way the answer is still written out token by token. \emph{Rather than deciding how much to
  think}, we explore a different paradigm, \emph{zero-or-$T$ thinking}: a single model
  either answers from the implicit thinking inside one forward pass, generating no token,
  or performs full autoregressive reasoning at a cost of $T$ tokens when that is worth
  paying. We realise it in a
  \emph{hybrid decision model} (HDM), where one frozen backbone serves both modes and a
  lightweight \emph{self-routing block} chooses between them. The block estimates the
  \emph{value of thinking}, the utility the slow mode is expected to add over the fast one,
  from what that implicit thinking has already produced, and the slow mode runs only where
  that value exceeds a price $\lambda$ set per request.
  Where no per-decision label exists, as in episodic agentic tasks, we introduce
  \emph{Calibrated Decision Policy Optimisation} (CDPO), a reinforcement-learning
  recipe that trains the block from episode-level returns.
  We evaluate HDM on MMLU-Pro, AGIEval, GPQA-Diamond, JevBench, BANKING77 and
  MiniWoB. Specifically, on JevBench HDM-4B scores $0.857$
  against Jev's $0.851$ ($+0.7\%$) at a median $0.179$\,s against $0.652$
  ($-72.5\%$). On the agentic venue MiniWoB HDM-27B cuts token cost by
  $80.1\%$.
  \end{abstract}

\section{Introduction}

In Kahneman's dual-process theory of cognition, ``System~1 operates automatically and quickly, with little or no effort and no sense of voluntary control,'' whereas ``System~2 allocates attention to the effortful mental activities that demand it'' \citep{kahneman2011thinking}. This division has shaped the design philosophy of reasoning in large language models (LLMs) \citep{wang2025tutorial}, but the work it has influenced builds either a good System Two or a good System One, and a model that is both remains underexplored.

\begin{figure}[!t]
\centering
\begin{minipage}[t]{0.49\columnwidth}\vspace{0pt}\centering
\includegraphics[width=\linewidth]{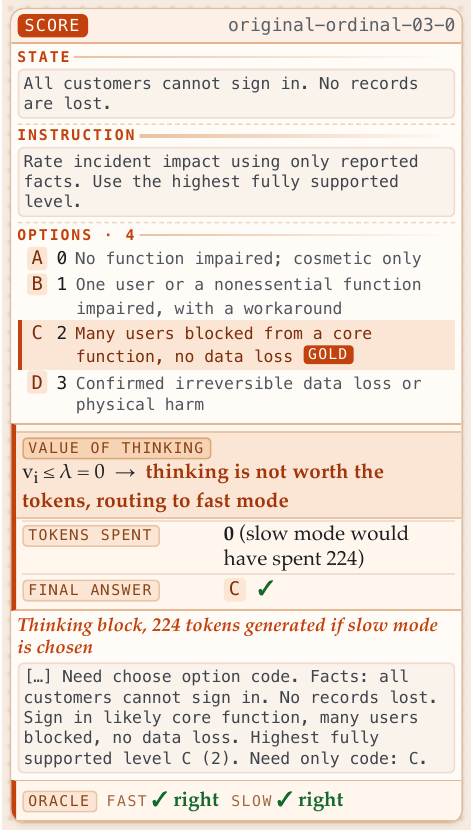}
\end{minipage}\hfill
\begin{minipage}[t]{0.49\columnwidth}\vspace{0pt}\centering
\includegraphics[width=\linewidth]{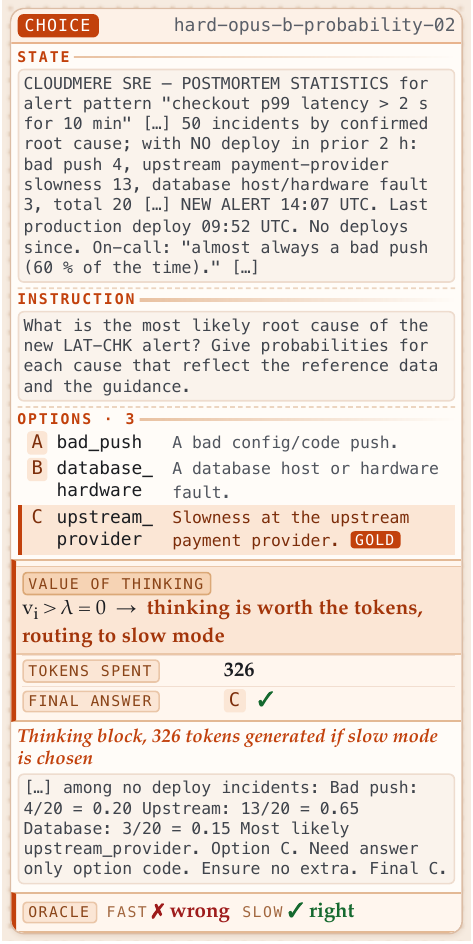}
\end{minipage}
\caption{\textbf{HDM's zero-or-$T$ thinking on two JevBench questions.} LEFT, a
case where the fast mode is sufficient: the predicted value of thinking is below
$\lambda$, the block answers at no token, and running both modes
afterwards confirms the slow mode would not have changed the answer. RIGHT, a case that should go to the slow
mode: the predicted value is above $\lambda$, routing to the slow mode, and running both
modes afterwards confirms the fast answer would have been wrong.
Appendix~\ref{app:cases} gives both thinking blocks in full.}
\label{fig:teaser}
\end{figure}

Most current System Two models (like reasoning models) raise their capability through autoregressive test-time scaling, including OpenAI's GPT-6, Anthropic's Claude Fable~5 and the open-weight Qwen3.8, which write a chain of thought before the answer, at a cost in tokens and latency on every question. A recent class of System One models, Jev \citep{typesafe2026jev} and AnyJev \citep{nokia2026anyjev} among them, takes the strict case, emitting no tokens at inference and producing the answer in a single non-autoregressive pass, each adapting a reasoning-model backbone rather than training a different kind of model.

\begin{figure*}[t]
\centering
% Scaled at inclusion rather than redrawn: the figure's sources are another
% author's working copy, and \textwidth is what the full-width float allows.
\includegraphics[width=0.82\textwidth]{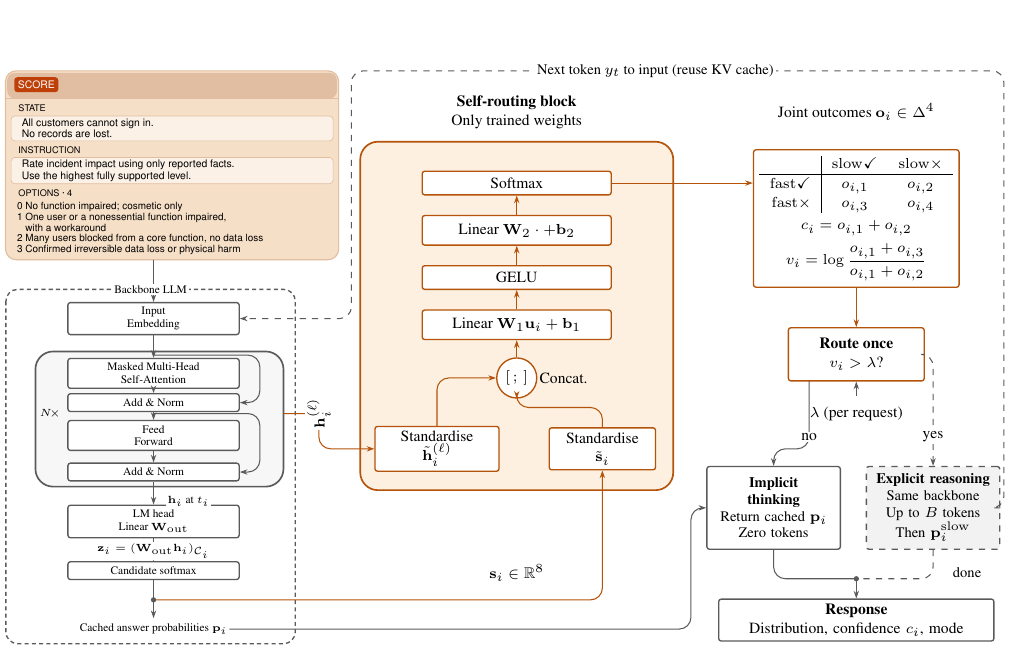}
\caption{\textbf{Backbone sketch of HDM.} The frozen backbone encodes a typed request with no token generated; the self-routing block, which holds the only learned weights, reads the readout state $\bm{h}_i$ and eight statistics of the fast answer $\bm{p}_i$ and returns the joint outcome of the two modes, from which $c_i$ and $v_i$ follow. Routing occurs once: the model either returns the cached fast answer or decodes autoregressively with the same backbone. The dashed feedback arrow returns each generated token to the shared context, reusing the KV cache; only the completed answer enters the response.}
\label{fig:arch}
\end{figure*}

One line of work seeks efficiency inside reasoning models themselves, shortening the chain of thought without losing accuracy \citep{snell2024scaling,muennighoff2025s1,zhang2025tokensqueeze,aggarwal2025l1,zhang2025adaptthink}. These methods make the chain cheaper rather than optional: they ask how long the reasoning should run, not whether a given question needs it at all. Another relieves the cost by routing between models, or down a cascade until an answer is accepted \citep{chen2023frugalgpt,ong2024routellm}. Every branch is a model generating an answer, so the choice on offer is between an expensive answer and a cheaper one.

\emph{Can one model keep both advantages and switch between them when it should, generating nothing at all where that suffices and reasoning where it does not?} In this work, we introduce \emph{zero-or-$T$ thinking}, an adaptive-inference paradigm that unifies token-free fast decision with full autoregressive reasoning inside a single language model: \emph{rather than deciding how much to think, the model decides whether autoregressive thinking is needed at all}, and a question costs either no generated token or the $T$ a reasoning pass takes. Specifically, we define the \emph{value of thinking} on a question as the utility the slow-thinking mode is expected to add over the fast-thinking one, and a question is escalated only when that value exceeds a price $\lambda$ set on the request.

We instantiate zero-or-$T$ thinking in a \emph{hybrid decision model} (HDM): a fast-and-slow model, its \emph{fast-thinking} mode producing a decision directly from the implicit thinking with no token generated, its \emph{slow-thinking} mode the backbone's own reasoning inherited unchanged, and a learned \emph{self-routing block} choosing between them before any token is generated. HDM estimates the conditional gain of thinking
from hidden representations the fast mode has already computed, and completes
cost-aware routing before inference begins.

We also address how the block is trained. On single-turn questions a gold answer is available, so running both modes constructs the better choice directly and supervision suffices; in episodic tasks the return is decided by many successive actions and no per-decision label exists, so \emph{we introduce a PPO-based reinforcement-learning recipe for training HDM, Calibrated Decision Policy Optimisation (CDPO).} CDPO maximises the return an episode earns discounted by the escalations it spent, with that one episode-level signal shared by every decision inside it.

We evaluate HDM on MMLU-Pro, AGIEval, GPQA-Diamond, JevBench, BANKING77 and MiniWoB. On JevBench HDM-4B reaches $0.857$ against Jev's
$0.851$ at a median $0.179$\,s against $0.652$; on AGIEval HDM-27B
trades $1.5$ accuracy points for $37\%$ fewer thinking tokens, and on MiniWoB $5$
points of reward for $80\%$ fewer.

Our contribution is threefold. \textit{First}, we propose \textit{zero-or-$T$ thinking}, a paradigm in which a question costs either no generated token or the $T$ a reasoning pass takes and one frozen backbone serves both modes, switched by a self-routing block of a few hundred kilobytes that reads representations the fast pass has already computed, so the choice costs no forward pass and no token. \textit{Second}, we introduce \textit{the value of thinking} as the criterion that block switches on: the utility the slow-thinking mode is expected to add over the fast-thinking one, one interpretable quantity thresholded against a price $\lambda$ set on the request rather than a number tuned per benchmark. \textit{Last}, we introduce \textit{CDPO}, which trains the block where no per-decision label exists, from the return an episode earns discounted by the escalations it spent.

\section{Method}
\label{sec:method}

A \emph{hybrid decision model} is a model $f$ with two decision modes, $f_{\mathrm{fast}}$ and $f_{\mathrm{slow}}$, and a policy $\pi$ that picks one of them for each question $q$, which is then answered by $f_{\pi(q)}$,
\begin{equation}
  \pi(q)\in\{\mathrm{fast},\mathrm{slow}\},
  \label{eq:hybrid}
\end{equation}
and is \emph{self-routing} when $\pi$ reads only what $f$'s implicit thinking has already produced.

In this paper we assume $f$ is a transformer-based large language model. The slow mode $f_{\mathrm{slow}}$ (or \emph{explicit thinking}) generates tokens as normal reasoning LLMs do. The fast mode $f_{\mathrm{fast}}$ (or \emph{implicit thinking}) generates no tokens, producing the decision by multi-class classification or numerical prediction, depending on the task type \citep{devlin2019bert}. Switching between the two modes lets a model reason where reasoning helps and respond rapidly where there is no need for thinking, while preserving the advantage of the world knowledge stored in the LLM.

\emph{Predicting the value of thinking is the key to knowing when to switch.} Confidence is no substitute for it: confidence reports how sure the fast answer is, and that is a different question from whether the slow mode would answer differently. The two come apart in both directions: thinking sometimes spoils an answer the fast mode had right, and it adds nothing where both modes would answer correctly. A router therefore needs the two outcomes jointly.

\paragraph{Stating the routing problem as an objective.} We define the \emph{utility} a mode earns on question $i$ as its correctness,
\begin{equation}
  u_i(m)=a_i^{m}=\mathbf{1}\big[\text{mode $m$ answers $i$ correctly}\big]\in\{0,1\},
  \label{eq:utility}
\end{equation}
for $m\in\{\mathrm{fast},\mathrm{slow}\}$, so running both modes observes one of four outcomes: both right, the fast one only, the slow one only, or neither. A correct answer is worth less the longer it takes, and worth nothing once it is too late, so the wait discounts the utility rather than subtracting from it. The \emph{reward} a mode earns is therefore its utility under an exponential discount on the tokens it generates, at a rate we call the \emph{price of thinking} $\lambda$,
\begin{equation}
  r_i(m,\rho)=u_i(m)\,e^{-\rho T_i^{m}},
  \label{eq:reward}
\end{equation}
Each generated token multiplies the utility by $e^{-\rho}$. A wrong answer is worth zero at every price, and the slow mode runs on question $i$ only when the value of thinking there exceeds the price, $v_i>\lambda$~(\ref{eq:escalate}), that is, only when thinking raises the chance of a correct answer by a factor of at least $e^{\lambda}$. With $T_i^{\mathrm{fast}},T_i^{\mathrm{slow}}$ the tokens each generates ($T_i^{\mathrm{fast}}=0$ in HDM), a router $\pi$ attains on a set $\mathcal{D}$ a mean utility $\bar u$ and a mean token cost $\bar T$,
\begin{equation}
  \bar u(\pi)=\frac{1}{|\mathcal{D}|}\sum_{i\in\mathcal{D}} u_i(\pi(i)),
  \qquad
  \bar T(\pi)=\frac{1}{|\mathcal{D}|}\sum_{i\in\mathcal{D}} T_i^{\pi(i)} .
  \label{eq:objective}
\end{equation}
We want the router maximising $\bar u$ at a given $\bar T$ under two constraints: neither $a_i^{\mathrm{fast}}$ nor $a_i^{\mathrm{slow}}$ has been obtained when it decides, and it reads only quantities $f$ has already computed, so it adds no forward pass and no token.

\paragraph{Typed input schema.} We define a \emph{typed general request} to carry a \emph{prompt} $s$ (text or multimodal media), optional explicit \emph{questions} $q_1,\dots,q_N$, and an optional \emph{candidate answer set} for each; only the prompt is required, so the schema names structure an ordinary request already carries and a bare prompt is already a free-form request. The schema makes the two modes substitutable, since a caller receives an answer of the same type whichever produced it, and it lets HDM serve a choice among candidates, a binary judgement, a graded score and free-form generation with one backbone and one block.

\subsection{The hybrid decision model}
\label{sec:htm}

HDM is a frozen $f$ with one learned block added (Figure~\ref{fig:arch}). The policy $\pi$ has two parts: the \emph{self-routing block}, a small network on the hidden state $\bm{h}_i$ of one of the backbone's transformer layers, and the \emph{price of thinking} $\lambda$, which the user sets per request. One trained block covers the whole cost--accuracy curve, so varying $\lambda$ moves the operating point along that curve without retraining and without a second block.

\subsubsection{Architecture}
\label{sec:arch}

Given a typed input request as described above, three components run in order (Figure~\ref{fig:arch}). HDM first encodes the prompt with the fast-thinking mode and reads an answer $\bm{p}_i$ at the readout position, producing on the same pass the hidden state $\bm{h}_i$. The self-routing block then reads $\bm{h}_i$ and eight statistics of $\bm{p}_i$ (Table~\ref{tab:hyper} lists them), and returns a distribution $\bm{o}_i$ over the four joint values of equation~\ref{eq:utility}'s utility pair $(a_i^{\mathrm{fast}},a_i^{\mathrm{slow}})$, so $\bm{o}_i$ estimates both modes' utility before either has run. Finally, the slow-thinking mode runs only on escalated questions and continues the generation autoregressively from the state the fast pass has already computed.

The block's $\bm{o}_i$ has two marginals, each estimating a utility in equation~\ref{eq:utility}: the confidence reported with the fast answer, $c_i=\bm{o}_{i,1}+\bm{o}_{i,2}=P(\mathrm{fast}\checkmark)$, for $\mathbb{E}[a_i^{\mathrm{fast}}]$; and the \emph{value of thinking}, which equation~\ref{eq:reward} makes the log of the ratio the two modes succeed at,
\begin{equation}
  v_i=\log\frac{P(\mathrm{slow}\checkmark)}{P(\mathrm{fast}\checkmark)}
         =\log\frac{\bm{o}_{i,1}+\bm{o}_{i,3}}{\bm{o}_{i,1}+\bm{o}_{i,2}},
  \label{eq:gain}
\end{equation}
rather than the difference a linear price gives, which shrinks where thinking is worth most.

\label{sec:fast}

\paragraph{Fast-thinking mode.} With the candidates enumerated, HDM's fast-thinking mode emits no tokens. Let $x_i$ be the prompt for question $i$, which includes the candidates with their codes and ends in a boundary string that makes the answer the next thing $f$ would produce, and let $t_i=|x_i|$ be its last position, the \emph{readout position}. One forward pass of $f$ over $x_i$ leaves a hidden state $\bm{h}_i\in\mathbb{R}^d$ at $t_i$, and
\begin{equation}
\begin{split}
  \bm{z}_i&=\big(\bm{W}_{\mathrm{out}}\,\bm{h}_i\big)_{\mathcal{C}_i}\in\mathbb{R}^{K_i},\\
  \bm{p}_i&=\mathrm{softmax}(\bm{z}_i),
  \end{split}
  \label{eq:fast}
\end{equation}
where $\bm{W}_{\mathrm{out}}\in\mathbb{R}^{|V|\times d}$ is $f$'s output projection onto its vocabulary $V$, giving one logit per token, and $\mathcal{C}_i$ indexes the codes of question $i$'s $K_i$ candidates, so the candidate logits $\bm{z}_i$ score the candidates against one another. We also implement a readout scoring each candidate by its own text, but for uniformity every experiment here reads the code index.

\label{sec:slow}

\paragraph{Slow-thinking mode.} Question $i$ is escalated when the value of thinking exceeds the price, $v_i>\lambda$~(\ref{eq:escalate}). On escalation the first token is generated from the prefix the fast pass has already encoded, as usual, and decoding proceeds autoregressively up to a thinking budget of $B$ tokens, and the self-routing block is not consulted again, so one decision is taken per question rather than per token. Both modes are rendered from one prompt, so the context is shared and an escalation pays no second prefill. Appendix~\ref{app:prompts} gives the template mechanics.

\subsubsection{The price of thinking}
\label{sec:routing}

Without a shared budget the expected reward equation~\ref{eq:reward} decomposes across questions, so each question is decided on its own. The fast mode generates nothing, so its reward is $P(\mathrm{fast}\checkmark)$, against the slow mode's $P(\mathrm{slow}\checkmark)\,e^{-\rho T_i^{\mathrm{slow}}}$, and escalating pays exactly when $v_i>\rho\,T_i^{\mathrm{slow}}$. That right-hand side is a property of the question rather than of the request: it is not known until the slow pass has run. We therefore price a slow pass at its mean length $\bar{T}^{\mathrm{slow}}$ and route on the \emph{price per escalation} $\lambda=\rho\,\bar{T}^{\mathrm{slow}}$, which settles the trade-off between the value of thinking $v_i$~(\ref{eq:gain}) and the price with one threshold:
\begin{equation}
  \text{escalate}_i=\mathbf{1}[v_i>\lambda].
  \label{eq:escalate}
\end{equation}
Written as a mode, this is the policy $\pi$ of equation~\ref{eq:hybrid}: escalating
means $\pi(q_i)=\mathrm{slow}$.
Escalating on $\lambda$ approximates the exact rule $v_i>\rho\,T_i^{\mathrm{slow}}$: charging every escalation alike misprices question $i$ by $\rho(T_i^{\mathrm{slow}}-\bar{T}^{\mathrm{slow}})$, so it escalates too readily where thinking would run long and too rarely where it would be brief. Everywhere else in this paper, including Table~\ref{tab:results}'s rows and the prices CDPO draws, $\lambda$ is the price per escalation; $\rho$ is equation~\ref{eq:reward}'s price per token and is used nowhere else.

\subsubsection{What does thinking actually buy?}
\label{sec:causal}

This section sets out why a decoder-only backbone can serve as an encoder. Generation is not all a decoder-only forward pass does. Before any token is
produced, the backbone encodes the request: tokens enter a latent space and each
layer abstracts what the layer below it holds. \emph{This is thinking the model does without
writing it down}, studied as latent or implicit reasoning
\citep{yang2024latenthop,wang2024grokked,hao2024coconut}, and the state it
leaves is information the model has already worked over rather than the input in
another form.

Our own experiments carry evidence for this. In
Table~\ref{tab:jevbench} the fast mode answers most items without generating. In
Table~\ref{tab:features} the same state tells the routing block whether thinking
would change that answer, and tells it more than the fast answer's own output
distribution does.

Thinking buys \emph{steps of deliberation}, each generated token re-entering
the context so that the next pass starts
further along \citep{li2024cotserial}. Hence the value of
thinking, our estimate of where more of that deliberation would change the
answer.

\subsection{Calibrated decision policy optimisation for episodic tasks}
\label{sec:pmpo}

\emph{In a multi-turn agentic task a single question's correctness usually does not decide whether the task succeeds, and often no per-question label exists to train against.} (Where a label can be obtained, we can train the same block by self-labelled supervision, Section~\ref{sec:training}.)

We therefore introduce \emph{Calibrated Decision Policy Optimisation} (CDPO), a PPO-based reinforcement-learning recipe that learns from the episode's return. We first fix the notation. Write $e_t\in\{0,1\}$ for whether step $t$ escalated, and $R$ for the terminal return, which is equation~\ref{eq:utility}'s utility at episode granularity. A price $\lambda_t$ is drawn for each decision. The objective is that return under the discount equation~\ref{eq:reward} puts on one question, $G_t=R\,e^{-\lambda_t\sum_{s} e_s}$. The exponent counts escalations where equation~\ref{eq:reward} charges tokens, by the normalisation of Section~\ref{sec:routing}: an escalated step reasons to its cap and the fast mode generates nothing, so the per-step difference $T_t^{\mathrm{slow}}-T_t^{\mathrm{fast}}$ is about constant and absorbs into $\lambda$, which is again a price per escalation. The block is trained on a soft form of the rule, $\pi_\theta(e_t\mid x_t,\lambda_t)=\sigma\big((v_t-\lambda_t)/\tau\big)$ at temperature $\tau$, which is differentiable in $\theta$ and recovers equation~\ref{eq:escalate} as $\tau\to0$. Episodes are collected under the policy $\mu$ that was acting, so each recorded decision is reweighted by $\rho_t=\pi_\theta/\mu$. Ratios multiplied along tens of steps collapse, so the ratio is taken per decision inside the clipped surrogate of PPO \citep{schulman2017ppo}, which makes CDPO a surrogate for that objective rather than an unbiased estimator of its gradient (Appendix~\ref{app:pmpo}), with the advantage $A_t=G_t-\bar G$, whose return is the episode's and whose baseline $\bar G$ is the mean return at the same drawn prices:
\begin{equation}
  \mathcal{L}(\theta)=-\,\mathbb{E}_{t,\lambda}\Big[\min\big(\rho_t A_t,\ \mathrm{clip}(\rho_t,1-\epsilon,1+\epsilon)\,A_t\big)\Big].
  \label{eq:cdpo}
\end{equation}

\subsection{Self-labelled supervised training for non-episodic tasks}
\label{sec:training}

\begin{table*}[t]
\caption{\textbf{Evaluating how far HDM compresses token and wall-clock cost on general-purpose question answering.} Accuracy, escalation and cost at five prices. Escalation is the share of questions sent to the slow-thinking mode; tokens and seconds are per question. All six columns are at $B=4096$, layer $-1$, one question to a request. Brackets give a cost's change against that block's \emph{Slow only} row, and \textbf{bold} marks our rows. $\lambda$ prices one escalation, so it is comparable down a column; a slow pass costs a different number of tokens on each benchmark and on each backbone, and the same $\lambda$ is a different price across either.}
\label{tab:results}
\centering
\scriptsize
\setlength{\tabcolsep}{2.5pt}
\begin{tabular}{ll rrrr rrrr rrrr}
\toprule
& & \multicolumn{4}{c}{MMLU-Pro} & \multicolumn{4}{c}{AGIEval} & \multicolumn{4}{c}{GPQA-Diamond} \\
\cmidrule(lr){3-6}\cmidrule(lr){7-10}\cmidrule(lr){11-14}
Policy & $\lambda$ & Esc. & Acc. & Tok. & Sec. & Esc. & Acc. & Tok. & Sec. & Esc. & Acc. & Tok. & Sec. \\
\midrule
\multicolumn{14}{l}{\emph{Qwen3.5-4B}} \\
Fast only & -- & 0\% & 0.543 & 0 {\tiny($-$100\%)} & 0.27 {\tiny($-$98\%)} & 0\% & 0.606 & 0 {\tiny($-$100\%)} & 0.31 {\tiny($-$98\%)} & 0\% & 0.429 & 0 {\tiny($-$100\%)} & 0.27 {\tiny($-$99\%)} \\
Slow only & -- & 100\% & 0.853 & 2281 & 12.30 & 100\% & 0.842 & 2364 & 12.79 & 100\% & 0.717 & 3685 & 19.75 \\
\textbf{HDM} & $-0.1$ & 97\% & 0.844 & 2220 {\tiny($-$2.7\%)} & 11.93 {\tiny($-$3.0\%)} & 100\% & 0.842 & 2364 {\tiny(0\%)} & 12.75 {\tiny($-$0.4\%)} & 100\% & 0.717 & 3685 {\tiny(0\%)} & 19.56 {\tiny($-$1.0\%)} \\
\textbf{HDM} & $0$ & 66\% & 0.844 & 1845 {\tiny($-$19\%)} & 9.92 {\tiny($-$19\%)} & 71\% & 0.830 & 1801 {\tiny($-$24\%)} & 9.75 {\tiny($-$24\%)} & 93\% & 0.707 & 3491 {\tiny($-$5.3\%)} & 18.58 {\tiny($-$5.9\%)} \\
\textbf{HDM} & $0.05$ & 58\% & 0.835 & 1720 {\tiny($-$25\%)} & 9.27 {\tiny($-$25\%)} & 57\% & 0.827 & 1465 {\tiny($-$38\%)} & 7.98 {\tiny($-$38\%)} & 82\% & 0.687 & 3089 {\tiny($-$16\%)} & 16.47 {\tiny($-$17\%)} \\
\textbf{HDM} & $0.1$ & 53\% & 0.815 & 1574 {\tiny($-$31\%)} & 8.53 {\tiny($-$31\%)} & 53\% & 0.818 & 1381 {\tiny($-$42\%)} & 7.55 {\tiny($-$41\%)} & 78\% & 0.687 & 2962 {\tiny($-$20\%)} & 15.85 {\tiny($-$20\%)} \\
\textbf{HDM} & $0.5$ & 39\% & 0.772 & 1166 {\tiny($-$49\%)} & 6.38 {\tiny($-$48\%)} & 39\% & 0.797 & 991 {\tiny($-$58\%)} & 5.50 {\tiny($-$57\%)} & 52\% & 0.621 & 1947 {\tiny($-$47\%)} & 10.47 {\tiny($-$47\%)} \\
\midrule
\multicolumn{14}{l}{\emph{Qwen3.8-27B}} \\
Fast only & -- & 0\% & 0.650 & 0 {\tiny($-$100\%)} & 0.46 {\tiny($-$97\%)} & 0\% & 0.710 & 0 {\tiny($-$100\%)} & 0.57 {\tiny($-$97\%)} & 0\% & 0.510 & 0 {\tiny($-$100\%)} & 0.47 {\tiny($-$99\%)} \\
Slow only & -- & 100\% & 0.902 & 778 & 17.07 & 100\% & 0.925 & 824 & 18.14 & 100\% & 0.848 & 2571 & 54.85 \\
\textbf{HDM} & $-0.1$ & 99\% & 0.902 & 770 {\tiny($-$1.0\%)} & 16.83 {\tiny($-$1.4\%)} & 99\% & 0.925 & 815 {\tiny($-$1.1\%)} & 17.84 {\tiny($-$1.6\%)} & 100\% & 0.848 & 2571 {\tiny(0\%)} & 54.79 {\tiny($-$0.1\%)} \\
\textbf{HDM} & $0$ & 63\% & 0.896 & 664 {\tiny($-$15\%)} & 14.53 {\tiny($-$15\%)} & 79\% & 0.919 & 725 {\tiny($-$12\%)} & 15.93 {\tiny($-$12\%)} & 96\% & 0.848 & 2552 {\tiny($-$0.7\%)} & 54.35 {\tiny($-$0.9\%)} \\
\textbf{HDM} & $0.05$ & 48\% & 0.890 & 606 {\tiny($-$22\%)} & 13.32 {\tiny($-$22\%)} & 53\% & 0.916 & 579 {\tiny($-$30\%)} & 12.85 {\tiny($-$29\%)} & 82\% & 0.843 & 2324 {\tiny($-$9.6\%)} & 49.56 {\tiny($-$9.6\%)} \\
\textbf{HDM} & $0.1$ & 47\% & 0.887 & 598 {\tiny($-$23\%)} & 13.13 {\tiny($-$23\%)} & 50\% & 0.910 & 519 {\tiny($-$37\%)} & 11.54 {\tiny($-$36\%)} & 78\% & 0.828 & 2231 {\tiny($-$13\%)} & 47.59 {\tiny($-$13\%)} \\
\textbf{HDM} & $0.5$ & 38\% & 0.873 & 486 {\tiny($-$38\%)} & 10.60 {\tiny($-$38\%)} & 40\% & 0.890 & 439 {\tiny($-$47\%)} & 9.85 {\tiny($-$46\%)} & 55\% & 0.788 & 1463 {\tiny($-$43\%)} & 31.17 {\tiny($-$43\%)} \\
\bottomrule\end{tabular}
\end{table*}

For tasks where a correct label is easy to acquire, we can also train the self-routing block with a supervised objective, by treating the decision as a four-way classification with cross-entropy. At temperature zero the frozen backbone can be treated as a deterministic map, so the slow-thinking mode's correctness on a labelled question is fixed. Given a set of questions with gold answers, we run the fast-thinking mode (recording $\bm{h}_i^{(\ell)}$ for a few candidate layers $\ell$ and the statistics $\bm{s}_i$) and the slow-thinking mode (recording $a_i^{\mathrm{slow}}$ and $T_i$), compare both with gold, and obtain the four-way label $y_i\in\{0,1,2,3\}$ (Appendix~\ref{app:method}). Given $y_i$, the block is fitted on $\bm{h}_i$ and $\bm{s}_i$ by cross-entropy over the four classes,
\begin{equation}
  \mathcal{L}(\theta)=-\frac{1}{|\mathcal{D}|}\sum_{i\in\mathcal{D}}
  \log P_\theta\big(y_i\mid\bm{h}_i,\bm{s}_i\big),
  \label{eq:ce}
\end{equation}
where $P_\theta(\cdot\mid\bm{h}_i,\bm{s}_i)$ is the distribution $\bm{o}_i$ the block returns. No price appears in the loss, so one fit serves every $\lambda$.

% ---------------------------------------------------------------------------
% PENDING RE-MEASUREMENT, and the prose is deliberately ahead of the tables on
% the user's decision; the values are being replaced table by table.
% Two things changed under them. (1) The prompt: 9122c1d/76264fd unified the two
% renders (2026-10-09) -- one system turn for both modes, one render with
% thinking on, the fast mode closing the block. (2) The budget: B is now 4096,
% where every number in Tables 1-7 was formed at 1024, which bound hard -- 90.3%
% of MMLU-Pro questions reached that cap and the median sat exactly on it, so
% the published slow columns are lower bounds. A table's own caption states the
% budget its numbers were measured at until they are replaced.
% What moved: the fast prompt's system text, the question's position and the
% boundary's position, so every fast-mode readout is conditioned on a different
% string. The ALFWorld run in flight (29636, snapshot 42b7b4f) is already on the
% new design and its numbers will not be comparable with these rows.
% ---------------------------------------------------------------------------

\section{Experiments}

\label{sec:experiments}

\subsection{Setup}
\label{sec:setup}

\paragraph{Backbone.} We run every experiment on two backbones, Qwen3.5-4B and Qwen3.8-27B, the larger one repeating the main result at scale. We set the thinking budget to $B=4096$ tokens and decode the slow-thinking mode greedily. Under that cap it is truncated on $36.1\%$ of Qwen3.5-4B's questions and $14.9\%$ of Qwen3.8-27B's, whose token counts and accuracies are therefore lower bounds. We serve both backbones with SGLang; Appendix~\ref{app:setup} gives the request schema and that serving path, and Appendix~\ref{app:hyper} the hyperparameters.%
\footnote{Our hybrid decision framework is open source at
\ificmlshowauthors\url{https://github.com/zhengxuyu/HDM}%
\else\url{https://github.com/xxxx/xxxx}\fi.}

\paragraph{Data.} We build the training set from eleven of MMLU-Pro's \citep{wang2024mmlupro} fourteen subject categories and keep three as a held-out test set, so every number measures transfer to unseen subjects. All $1{,}500$ questions are drawn from MMLU-Pro's own test split, at stride 8 across all fourteen categories: chemistry, economics and psychology ($346$ questions) are the three held out, the other eleven ($1{,}154$) form the training set and the $K=4$ selection folds, and a deterministic hash of the category names picks them, so the split is reproducible rather than chosen. \emph{Holding out whole categories is legitimate here because that set calibrates an estimator rather than teaching a task}: what it has to span is the distribution of $(\bm{h}_i,\bm{s}_i)$, not the subject matter. Transfer is measured further out still, on every seventh question of AGIEval's English multiple-choice tasks at a pinned commit, $335$ of $2{,}340$, and on all $198$ of GPQA-Diamond. For agentic tasks we evaluate HDM on ALFWorld ($134$ held-out \texttt{valid\_unseen} games) and MiniWoB ($27$ held-out \texttt{click} tasks).

\paragraph{Readout layer.}
In our experiments we use the final layer as the readout layer throughout, because every transformer-based model has one. Our preliminary experiments found that the final layer is not always the best performing one (Section~\ref{sec:features}), but we keep it for the generality of the model and the consistency of the experiments.

\paragraph{Metrics.}
In our experiments we use correctness as utility, so $\bar u$ of equation~\ref{eq:objective} is accuracy and $\bar T$ is tokens per question. What we measure is the trade-off between the two: the thinking tokens and the wall-clock a policy saves while holding accuracy, or while giving up a little of it. We report accuracy on the held-out categories against escalation rate, swept over $\lambda$, and expected thinking tokens per question. No single price is baked in: training draws $\lambda$ per example from $[-0.2,0.5]$, so one block serves any request-time price, and the prices above $0.5$ in Figure~\ref{fig:lambda}'s sweep are beyond the range it drew.

\paragraph{Baselines.} Every table compares HDM against the two endpoints, fast only and slow only, and against the state of the art on that benchmark where one exists: Jev \citep{typesafe2026jev} and the other published System One models on JevBench (Table~\ref{tab:jevbench}), and AnyJev \citep{nokia2026anyjev} on BANKING77 (Table~\ref{tab:banking}).

\subsection{Evaluation on non-episodic tasks}
\label{sec:main}

\begin{table}[t]
\caption{Accuracy and latency on JevBench v1's 231 public items, under the coded readout of Appendix~\ref{app:readout}. Our rows are HDM-4B (Qwen3.5-4B) and HDM-27B (Qwen3.8-27B) at $B=4096$, each with a block trained on MMLU-Pro and nothing retrained here. Escalation is the share of items sent to the slow-thinking mode and P50 the median wall-clock seconds per item.}
\label{tab:jevbench}
\centering
\scriptsize
\setlength{\tabcolsep}{4pt}
\begin{tabular}{lrrr}
\toprule
System & Esc. & Accuracy & P50 (s) \\
\midrule
Laya, 421M & --- & $0.584$ & --- \\
smalljev semantic-v9 & --- & $0.606$ & --- \\
jeff, 400M & --- & $0.628$ & --- \\
Open-Jev, 2B & --- & $0.645$ & --- \\
kev, 4B & --- & $0.662$ & --- \\
kev, 8B & --- & $0.714$ & --- \\
jev-local & --- & $0.749$ & --- \\
LLM2Jev & --- & $0.762$ & $0.048$ \\
LEV-2.6B, depth 4 \citep{song2026lev} & --- & $0.762$ & $0.200$ \\
AnyJev Tacit-9B, fast only & $0$\% & $0.823$ & --- \\
Jev 1.13.0 & --- & $0.851$ & $0.652$ \\
AnyJev Tacit-9B, adaptive & --- & $0.887$ & --- \\
\midrule
\textbf{HDM}-4B, fast only & $0$\% & $0.762$ & $0.226$ \\
\textbf{HDM}-4B, always thinking & $100$\% & $0.931$ & $3.86$ \\
\textbf{HDM}-27B, fast only & $0$\% & $0.879$ & $0.411$ \\
\textbf{HDM}-27B, always thinking & $100$\% & $0.991$ & $4.81$ \\
\midrule
\textbf{HDM}-4B, routed $\lambda=0$ & $30$\% & $0.857$ & $0.179$ \\
\textbf{HDM}-27B, routed $\lambda=0$ & $54$\% & $0.987$ & $2.49$ \\
\bottomrule
\end{tabular}
\end{table}

\begin{table}[t]
\caption{Accuracy and cost on BANKING77 as a 20-way intent task, 300 items, on Qwen3.8-27B at $B=4096$, with a block trained on MMLU-Pro and nothing retrained here.}
\label{tab:banking}
\centering
\small
\setlength{\tabcolsep}{4pt}
\begin{tabular}{lrrrr}
\toprule
Policy & Esc. & Acc. & Tok. & Sec. \\
\midrule
Tacit-9B + AnyJev & 6.0\% & 0.913 & 140.2 & 3.82 \\
\midrule
Qwen3.8-27B + AnyJev $\tau=1.0$ & 7.0\% & 0.923 & 33.6 & --- \\
Qwen3.8-27B + AnyJev $\tau=0.5$ & 3.3\% & 0.933 & 16.9 & --- \\
\midrule
Slow only & 100\% & 0.913 & 135.3 & 3.53 \\
Fast only & 0\% & 0.923 & 0.0 & 0.57 \\
\midrule
\textbf{HDM}, $\lambda=0$ & 6.7\% & 0.920 & 14.1 & 0.83 \\
\textbf{HDM}, $\lambda=0.05$ & 0.3\% & 0.920 & 1.1 & 0.56 \\
\bottomrule
\end{tabular}
\end{table}

% Superseded 2026-10-15. The AGIEval and GPQA-Diamond columns are now 29777 and
% 29778 (plus 29867/29868 at the two high prices), not the 28324/28545 this note
% described, and the condition it set has been met: MMLU-Pro was re-run onto the
% same block, so all three 27B columns are block-27b-sglang-29734 and all three 4B
% columns are block-4b-sglang-29662-a681033. Both read feature_layers (-1,),
% verified from the files on 2026-10-15, as does every other block in the paper.
% What still differs between tables is which collection a block was fitted on, not
% which layer it reads: Table 2's 4B row is block-4b-sglang-29662 (before the
% feature-scaling correction) and Table 3 is block-27b-lastlayer at B=1024.
% The only layer-24 and layer-48 blocks left are block-4b-unweighted and
% block-27b-unweighted, which no row of any table uses any more.

\begin{figure}[bt]
\centering
\includegraphics{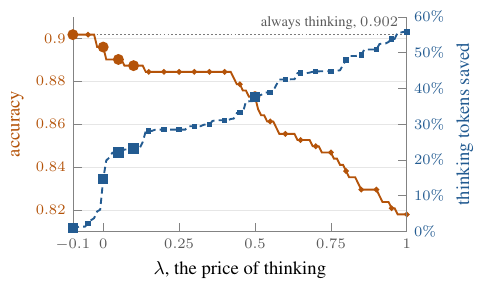}
\caption{\textbf{The price-of-thinking curve.} Qwen3.8-27B on MMLU-Pro's $346$ held-out questions. Solid line, left axis: accuracy. Dashed line, right axis: thinking tokens saved against always thinking. Dotted line: always thinking's accuracy, $0.902$. Each line passes through all $111$ prices swept at a step of $0.01$, marked at every fifth; the large marks are the five prices Table~\ref{tab:results} reports.}
\label{fig:lambda}
\end{figure}

We evaluate HDM on MMLU-Pro, AGIEval and GPQA-Diamond; Table~\ref{tab:results}
reports accuracy, escalation and cost at five prices on each, and
Figure~\ref{fig:lambda} sweeps the price behind one column. \emph{The
router buys almost all of what thinking is worth for a fraction of what it costs}:
at $\lambda=0.5$ Qwen3.8-27B gives up $0.029$ of always thinking's accuracy on
MMLU-Pro for $38\%$ fewer tokens, and $0.035$ on AGIEval for $47\%$ fewer. \emph{HDM retains the gain thinking
brings at a fraction of its cost}: $88\%$ of it on MMLU-Pro for $62\%$ of the
tokens, and $84\%$ on AGIEval for $53\%$, where that gain over the fast mode is $0.252$ and $0.215$ of
accuracy. One set of weights serves all three columns of each backbone with nothing
retrained, so AGIEval and GPQA-Diamond measure transfer rather than fitting.
\emph{The saving comes from which questions
are escalated, not how many}: at $\lambda=0.5$ the block thinks on $38\%$ of MMLU-Pro
and saves $38\%$ of the tokens, where questions of equal length would have saved
$62\%$, so it is escalating the long ones.

\begin{table*}[t]
\caption{\textbf{Evaluating CDPO-trained HDM on episodic agentic tasks, with one outcome an episode rather than one a decision.} Reward, escalation and cost per episode on two agentic venues, Qwen3.8-27B throughout. \emph{Esc.} is the share of steps sent to the slow-thinking mode, \emph{Reward} the environment's terminal signal, and brackets give the change against that venue's \emph{Slow only} row. ALFWorld is $134$ held-out \texttt{valid\_unseen} games at $256$ thinking tokens a step, one run at $\lambda=0$, its block fitted on one round of $150$ episodes collected under random routing; MiniWoB is $27$ held-out \texttt{click} tasks at $512$ a step, its routed row pooled over three seeds, and its endpoints one run each.}
\label{tab:agentic}
\centering
\scriptsize
\setlength{\tabcolsep}{4pt}
\begin{tabular}{l rrrr rrrr}
\toprule
& \multicolumn{4}{c}{ALFWorld} & \multicolumn{4}{c}{MiniWoB} \\
\cmidrule(lr){2-5}\cmidrule(lr){6-9}
Policy & Esc. & Reward & Tok./ep & Sec./ep & Esc. & Reward & Tok./ep & Sec./ep \\
\midrule
% MiniWoB columns are 29773. **This comment used to say "the re-run under
% eq:reward" and that is false**: app:miniwob states the run "was collected and
% trained before the reward became the discount of eq:reward. It priced an episode
% linearly, R - lambda * sum_t e_t, and escalated on the difference
% P(slow ok) - P(fast ok) rather than on the log ratio". exps verified the linear
% form against every recorded triple in policies.json on all five policies, and
% the caption says the same. The stale comment is not harmless -- it was read back
% as evidence that MiniWoB had been measured under the proposed objective, which
% would make the caption's caveat wrong rather than this line. Every routed cell is
% pooled over the three seeds' 81 episodes, which for equal episode counts gives
% the same answer as a mean of per-seed means for reward, tokens and seconds, and
% a different one for Esc.: pooling escalations over steps gives 0.263 and
% averaging the three rates gives 0.315. 0.263 is the consistent one -- it times
% 6.0123 steps returns the actual 1.5802 escalations an episode, where 0.315
% returns 1.895. Seed 2 is why: the highest rate over the fewest steps.
% policies.json's per-policy blocks are seed 0 alone and do not say so, which is
% where 0.1807 and 415.9 come from; do not read cost figures out of them.
Fast only & $0$\% & 0.463 & 0 {\tiny($-$100\%)} & 55.4 {\tiny($-$70\%)} & $0$\% & 0.407 & 0 {\tiny($-$100\%)} & 21.3 {\tiny($-$78\%)} \\
Slow only & $100$\% & 0.642 & 6254 & 185.2 & $100$\% & 0.519 & 3345 & 96.1 \\
\midrule
\textbf{HDM} & $46$\% & 0.560 & 3015 {\tiny($-$52\%)} & 108.4 {\tiny($-$41\%)} & $26$\% & 0.469 & 664 {\tiny($-$80\%)} & 31.3 {\tiny($-$67\%)} \\
\bottomrule
\end{tabular}
\end{table*}

\begin{table}[t]
\caption{\textbf{Evaluating the discriminative ability to filter out the questions thinking helps on, and which features supply it.} Qwen3.5-4B, the $1{,}500$-question MMLU-Pro selection, $B=1024$. Two AUROCs per configuration, in cross-validation over the 11 training subjects ($1{,}154$ questions) and on the three held-out categories ($346$): \emph{rescue}, whether thinking turns the fast answer from wrong to right, the \emph{routing AUROC}; and \emph{right}, whether the fast answer is already right. \emph{Dim} is the block's input width, hidden state plus eight statistics. The last row zeroes the statistics rather than dropping them. One seed.}
\label{tab:features}
\centering
\scriptsize
\setlength{\tabcolsep}{3pt}
\begin{tabular}{lrcccc}
\toprule
& & \multicolumn{2}{c}{CV (11 subjects)} & \multicolumn{2}{c}{Test (3 unseen)} \\
\cmidrule(lr){3-4}\cmidrule(lr){5-6}
Configuration & Dim & rescue & right & rescue & right \\
\midrule
Stats only & 8 & 0.671 & 0.763 & 0.701 & 0.840 \\
Layer $-1$, raw & 2568 & 0.765 & 0.758 & 0.716 & 0.821 \\
Layer $-1$, PCA-64 & 72 & 0.745 & 0.764 & 0.692 & 0.847 \\
Layer 16, raw & 2568 & 0.733 & 0.710 & 0.742 & 0.816 \\
Layer 16, PCA-64 & 72 & 0.719 & 0.737 & 0.726 & 0.845 \\
Layer 24, raw & 2568 & 0.778 & 0.764 & 0.762 & 0.827 \\
Layer 24, PCA-64 & 72 & 0.730 & 0.731 & 0.729 & 0.826 \\
Layer 24, PCA-64, no statistics & 64 & 0.728 & 0.655 & 0.708 & 0.758 \\
\bottomrule
\end{tabular}
\end{table}

\subsection{Evaluation on JevBench and BANKING77 against other System One models}

\label{sec:jevbench}

Table~\ref{tab:jevbench} gives the result on JevBench. At $\lambda=0$ HDM-4B
reaches $0.857$ against Jev's $0.851$ and answers a median item in $0.179$\,s against
$0.652$; HDM-27B reaches $0.987$ at $2.49$\,s. JevBench is also the one venue
that tests transfer across question \emph{types} as well as subjects --- its $231$
items are typed as the schema types them, $139$ \texttt{choice}, $74$ \texttt{noul}
(a yes-or-no judgement) and $18$ \texttt{score} (a graded one), where our other
benchmarks are multiple choice throughout --- and \emph{the block transfers to types
it never trained on}, escalating \texttt{noul} at $0.784$ against \texttt{choice}'s
$0.417$ at $\lambda=0$.

BANKING77, a 20-way intent task over $300$ items, is where the cost side is
clearest. Thinking does not pay here: always thinking reaches $0.913$ for $135.3$
tokens an item where the fast mode alone reaches $0.923$ for none, and the router
follows the cheaper mode, reaching $0.920$ at $\lambda=0.05$ for $1.1$ tokens an item.
\emph{Against the closest published System One model we score higher and spend two
orders of magnitude fewer tokens.} \citet{nokia2026anyjev} report no token cost, so we
ran their released checkpoint through their official codebase, under their adaptive
rule, on the identical $300$ items: Tacit-9B + AnyJev in Table~\ref{tab:banking}
escalates $6.0\%$ and scores $0.913$, at $140.2$ tokens an item against our $1.1$. At
$\lambda=0$ the router escalates at nearly the same rate, $6.7\%$, and scores $0.920$
for $14.1$.

\subsection{Evaluation of HDM on agentic benchmarks}
\label{sec:miniwob}

CDPO is introduced for the episodic setting, where the decisions inside one episode share a budget and a trajectory and the return arrives once at the end.

We evaluate HDM on ALFWorld and MiniWoB (Table~\ref{tab:agentic}), and the routing works at both. On ALFWorld the block at $\lambda=0$ reaches $0.560$ between a fast floor of $0.463$ and a ceiling of $0.642$, taking $54\%$ of the available gain for $48\%$ of the thinking tokens; on MiniWoB it reaches $0.469$ against always thinking's $0.519$, giving up $0.05$ of reward to cut thinking tokens by $80.1\%$ and wall-clock by $67.4\%$.

\subsection{Ablation study}
\label{sec:ablation}
% sec:features was \subsubsection{Feature representation and readout layer} until
% it became this subsection's only child, whose number carried no information.
% Both labels resolve to this subsection; the two \ref{sec:features} callers
% (body above, appendix's decision-block paragraph) read the same either way.
\label{sec:features}

We ablate the block's design: what it reads, from which layer, and through which bottleneck. Table~\ref{tab:features} scores each configuration by its \emph{routing AUROC}, how well $v_i$ ranks the questions thinking helps on, taken from the selection sweep's probes rather than from the trained block. The \emph{stats-only} variant there is the same block with the hidden state removed from its input, so it routes on the fast mode's output distribution alone. \emph{The hidden state is what carries the signal}: the best configuration, layer 24 read raw, reaches $0.778$ against that block's $0.671$ in cross-validation and $0.762$ against $0.701$ on held-out categories. Notably, \emph{the last layer is not the best of the three}: layer 24 leads it by $0.013$ of \emph{rescue} in cross-validation and $0.046$ on the held-out categories, on the one seed this table reports. We read the last layer everywhere else for generality, since every transformer has one and a layer two thirds of the way up does not transfer across backbones.

\begin{table}[t]
\caption{\textbf{Comparison with non-parametric baselines and prior
thinking-mode routing methods, in the thinking tokens needed to reach $90$\% of
always-thinking's accuracy; lower is better.} JevBench's $231$ public items,
content readout at $B=4096$, blocks fitted on MMLU-Pro's training categories and
transferred. Each cell is tokens per question, read off that rule's monotone
frontier.}
\label{tab:baselines}
\centering
\small
\begin{tabular}{lcc}
\toprule
Rule & Qwen3.8-27B & Qwen3.5-4B \\
\midrule
always thinking & $591$ & $1709$ \\
\midrule
random, $p$ swept & $213$ & $991$ \\
confidence & $138$ & $945$ \\
\midrule
ThinkSwitcher (reproduced) & $352$ & $1709$ \\
\mbox{RouteLLM} (reproduced) & $112$ & $822$ \\
\midrule
\textbf{HDM} & $103$ & $792$ \\
\bottomrule
\end{tabular}
\end{table}

\subsection{Comparison with direct routing baselines}
\label{sec:baselines}

We evaluate HDM's routing against four rules over the same two modes: a
confidence threshold on the fast mode, escalating at random, the switch of
\citet{liang2025thinkswitcher}, and \mbox{RouteLLM} \citep{ong2024routellm}. Neither learned rule publishes
code, so both are our reimplementations. \mbox{RouteLLM} routes between models,
so what we fit is its binary-classifier reduction to our mode pair, as
\citet{pan2025rtr} run it: it reads the same layer $-1$ state and eight
statistics our block reads, and is fitted on the same training categories, so
what the comparison isolates is the estimator.

Table~\ref{tab:baselines} fixes the accuracy and reads the cost.
\emph{Reaching $90$\% of always-thinking's accuracy costs us $103$ tokens a
question on the 27B}, against $112$ for \mbox{RouteLLM}, $138$ for a confidence
threshold and $591$ for always thinking; on the 4B we spend $792$ against
\mbox{RouteLLM}'s $822$. The ordering is not fixed, and depends on where the target is set. Appendix~\ref{app:cases} reads seven routed
items one by one.

\emph{The switch costs more than escalating at random}, $352$ against $213$ on
the 27B and the full budget on the 4B, so the questions it picks are worth less
than questions drawn at random.

\section{Related work}

\subsection{Explicit reasoning and its cost}
Reasoning models gain accuracy from a long chain of thought, and test-time scaling treats its length as a compute knob \citep{snell2024scaling,muennighoff2025s1}. Work on making that chain cheaper bounds its length \citep{aggarwal2025l1}, condenses traces \citep{zhang2025tokensqueeze} or allocates depth per token \citep{schuster2022calm}; all of it shortens thinking, and none lets the model answer without decoding. \citet{song2026lev} do answer without decoding, scaling a typed decision model by rerunning its own layers instead of generating, with a stopper that picks a depth per question from readouts the model has already produced. \emph{The knob is still how much.} Their stopper asks whether the current answer has stopped changing with depth, where ours asks whether reasoning would change it from wrong to right: \emph{rather than deciding how much to think, our model decides whether autoregressive thinking is needed at all}.

\subsection{Fast-thinking models and their readout}
A model is fast-thinking when it writes no reasoning down, reading the answer from a single pass. \emph{A single pass cannot add steps of deliberation}, whereas each generated token adds one \citep{li2024cotserial}, and our two modes sit at the ends of that trade-off. Reading an answer this way is the multi-class protocol of encoder classifiers and zero-shot multiple choice \citep{devlin2019bert,hendrycks2021mmlu,wang2024mmlupro}, which Jev \citep{typesafe2026jev} productised as a typed decision API with zero output tokens, since reproduced in the open \citep{ekzhang2026openjevsglang,zhihz2026openjev,mapika2026decider,nokia2026anyjev}. Token-free fast decision is therefore not itself new; what is new here is holding it and autoregressive reasoning in one frozen backbone and switching between them from state the model has already computed. Confidence estimation reads the same activations \citep{hendrycks2017baseline,geifman2017selective,kadavath2022know,azaria2023internal,alain2017probes}; we read them for a different target --- not whether the answer is right, but whether thinking would make it right.

\subsection{Hybrid thinking, mode switching and routing}
A third line lets one model choose per question: one family trains the backbone to emit a mode token \citep{fang2025thinkless,zhang2025adaptthink,lou2025adacot,jiang2025hrm}, another keeps it fixed and adds an external switch on predicted success rates \citep{liang2025thinkswitcher}, a pre-inference probe \citep{he2025selfroute} or perplexity \citep{lu2025car}, and a third routes between models rather than modes \citep{chen2023frugalgpt,ong2024routellm,ning2026hrbench}, among them the nearest-neighbour \mbox{KNN-Router} of \citet{pan2025rtr}. The closest is \citet{nokia2026anyjev}, escalating a typed decision on a fixed top-two margin rather than a learned predictor, and from ThinkSwitcher we inherit the frozen backbone, the self-produced labels and the thresholded margin. What separates HDM is that each of these computes its routing signal apart from the answer, and \emph{each of the modes they switch between still decodes its answer autoregressively}, so what they choose is how much reasoning to write rather than whether to decode at all. All are trained for likelihood where decision-focused learning would train for the decision \citep{elmachtoub2022spo,wilder2019melding}, and the quantity we predict already has three names: a value of computation in metareasoning \citep{horvitz1987reasoning,russell1991metareasoning,gershman2015computational}, a treatment effect in uplift modelling \citep{gutierrez2017uplift,kunzel2019metalearners}, a deferral decision in learning to defer \citep{mozannar2020defer}. \emph{Estimating it for two modes of one network, from a state that network has already produced, is what is new.}

\section{Conclusion}
\label{sec:conclusion}

HDM estimates the conditional gain of thinking
from hidden representations the fast mode has already computed, and completes
cost-aware routing before inference begins. The signal it reads is \emph{whether thinking would help} rather
than whether the fast answer is right, and the two differ wherever thinking
cannot rescue a question or would spoil one it had right. A few hundred
kilobytes of added parameters and no annotation beyond a labelled question set
are enough, and the price is set per request rather than at training time. What we
claim against published systems is competitiveness at a lower token cost and not
dominance: on BANKING77 two of the compared configurations answer more of the
items than we do, and the latencies we quote from other work were measured on
other backends. Where
no per-decision label exists, the CDPO we introduce trains the same block from
episode-level returns.

\bibliography{references}
\bibliographystyle{icml2026}

\clearpage
\appendix
\section{Experimental setup in full}
\label{app:setup}

\paragraph{Requests.}
We instantiate the request type with the typed-decision schema of Jev \citep{typesafe2026jev}, as published, and extend it only with an optional $\{\lambda,B\}$ block. A request carries a state $s$ and $N$ typed questions $q_1,\dots,q_N$; each question has a type in $\{\texttt{choice},\texttt{score},\texttt{noul}\}$ and a candidate set, namely the option keys for \texttt{choice}, the ordered levels for \texttt{score}, and $\{\text{false},\text{true}\}$ for \texttt{noul}. Question identifiers never enter the prompt and questions are isolated from one another.

\paragraph{Serving path.} Every number at $B=4096$ is served through SGLang, one question to a request. An escalation reuses the fast pass's prefill rather than paying a second one, which this path lets us measure instead of assert: over one run's slow readouts every one hit the cache, at a median of nearly the whole prompt, and almost all of them recomputed only the few tokens at the boundary.

\section{Hyperparameters}
\label{app:hyper}

Table~\ref{tab:hyper} collects the settings every block and every run uses, unless a section states otherwise. Nothing in it was tuned per benchmark: the transfer sets and the agentic venues are served by blocks fitted under these same values.

\begin{table*}[tp]
\caption{Settings used for every block and every run. Width $w$ is the one entry that follows the block's input rather than being fixed across configurations.}
\label{tab:hyper}
\centering
\small
\begin{tabular}{ll}
\toprule
Distribution statistics & max prob, norm.\ entropy, top-2 margin, concentration, $\log K$, type one-hot \\
Optimiser & AdamW, lr $10^{-3}$, weight decay 0.05, batch 256 \\
Regularisation & dropout 0.3, early stopping (patience 6) \\
CDPO & $\lambda\sim\mathcal{U}[-0.2,0.5]$, $\tau=0.05$, clip $\epsilon=0.2$, mean-return baseline; same optimiser, early stopping \\
Slow-mode decoding & greedy, $B=4096$, stop at \texttt{</think>} \\
Collection batch & $m=10$ questions a shared state \\
Agentic budget & $256$ tokens a step (ALFWorld), $512$ (MiniWoB) \\
\bottomrule
\end{tabular}
\end{table*}

\section{Prompt formats}
\label{app:prompts}

\paragraph{The shared prompt.}
System: \emph{``Evaluate the state using the question and labeled options. Answer with only the option code.''}, one system turn for both modes, taking no position on whether to reason, since the thinking block carries that distinction already. Saying \emph{``do not explain or reason aloud''} to one mode and \emph{``think it through first''} to the other is what forced two renders, and two renders diverge at the first token of the system turn. User: the state, a blank line, and \texttt{Question: \{json\}}, where the JSON carries the question type, the instructions and the options as \{code, key, description\} triples with single-token codes.

Both modes read their answer at \texttt{\textbackslash nAnswer:} from the same candidate codes, the fast-thinking mode immediately after the closed block and the slow-thinking mode after its reasoning; nothing is parsed from generated text in either.

\section{How the fast mode reads its answer}
\label{app:readout}

The fast mode needs a score per candidate, and we implement two readouts that
produce one.

The \emph{coded} readout maps each candidate to a single-token letter, writes the
letters into the prompt beside the candidates, and reads the model's
distribution over those letter tokens at one position. Every number elsewhere in
this paper was produced this way. Its weakness is structural: a code is a token,
a token carries a prior, and the prior attaches to whichever candidate occupies
that slot. On MMLU-Pro the positions this readout favours are measurably not where the
answers are, and the favoured position is not the same one on the two backbones.

The \emph{content} readout writes the candidates with no labels at all and scores
each one by its own text, teacher-forced after the same prefix. Nothing is
sampled; the cost is one prefill plus $k$ short continuations that share it. With
no label there is nothing for a token prior to attach to, though the candidates
are still \emph{listed} in an order, so a positional effect can survive and is
measured rather than assumed away.

A content score needs a normalisation, and three are natural. Writing $\ell_o$
for the summed log-probability of option $o$, $|o|$ for its token count, and
$\ell_o^{\varnothing}$ for its score after a question with nothing in it to
answer: \textsc{sum} ranks on $\ell_o$, \textsc{mean} on $\ell_o/|o|$, and
\textsc{pmi} on $\ell_o-\ell_o^{\varnothing}$, the last being the standard
correction for surface-form competition \citep{holtzman2021surface}. Only
\textsc{pmi} pays for the content-free pass.

\section{Self-labelled data collection}
\label{app:method}

Where the training data carries an immediate reward, the records for whether to enter the slow-thinking mode can be self-labelled rather than annotated, and the self-routing block fitted on them by supervision. Algorithm~\ref{alg:collect} is that construction.

With the four-way label $y_i$, the block is fitted by equation~\ref{eq:ce}'s cross-entropy over the
joint outcome: a record contributes the log probability the block gives the outcome that
occurred, and nothing else, so neither mode is supervised on its own.

\begin{algorithm}[t]
\caption{Self-labelling data collection}
\label{alg:collect}
\small
\begin{algorithmic}[1]
\REQUIRE frozen backbone $f$; questions $\{(s_i,q_i,y^\star_i)\}$ with gold answers, $m$ to a batch over one shared state $s$; readout layers $L$; thinking budget $B$
\FOR{each batch}
  \STATE $\{(\bm{z}_i,\bm{h}_i^{(\ell)})\}_{\ell\in L}\leftarrow\textsc{Fast}(f,s,\{q_i\})$ \COMMENT{candidate scores and readout hidden states, equation~\ref{eq:fast}; one prefill}
  \STATE $\{(\bm{z}_i^{\mathrm{slow}},T_i)\}\leftarrow\textsc{Slow}(f,s,\{q_i\},B)$ \COMMENT{thinking on to the same readout, $T_i$ tokens spent}
  \FOR{each question $i$}
    \STATE $a_i^{\mathrm{fast}}\leftarrow\mathbf{1}[\arg\max\bm{z}_i=y^\star_i]$, $a_i^{\mathrm{slow}}\leftarrow\mathbf{1}[\arg\max\bm{z}_i^{\mathrm{slow}}=y^\star_i]$ \COMMENT{equation~\ref{eq:utility}'s pair; the gold answer is the only label used}
    \STATE $y_i\leftarrow 2(1-a_i^{\mathrm{fast}})+(1-a_i^{\mathrm{slow}})\in\{0,1,2,3\}$\label{line:label} \COMMENT{both right, fast only, slow only, neither}
    \STATE $\bm{s}_i\leftarrow\textsc{Stats}(\mathrm{softmax}(\bm{z}_i),\mathrm{type}(q_i))$ \COMMENT{the statistics of Table~\ref{tab:hyper}}
    \STATE store $(\bm{h}_i^{(\ell)},\bm{s}_i,y_i,a_i^{\mathrm{fast}},a_i^{\mathrm{slow}},T_i,\mathrm{category}(q_i))$
  \ENDFOR
\ENDFOR
\end{algorithmic}
\end{algorithm}

\section{The calibrated decision policy optimisation algorithm}
\label{app:pmpo}

\paragraph{What the surrogate does and does not do.} CDPO optimises a surrogate, not
the episode-level objective itself, and three gaps separate them. The per-decision
importance ratio replaces the trajectory ratio, which is what keeps the weights usable
over tens of steps but biases the estimator in a direction the clip controls rather than
removes. Credit is not assigned: every decision in an episode is scored by that episode's
return, so a decision is rewarded for what the decisions after it did, and with tens of
decisions an episode this is the weakest assumption in the recipe. And the prices are
redrawn each epoch while the trajectories are not, so the policy is re-scored at prices
the behaviour policy never acted under. We report the effective sample size
$(\sum_t\rho_t)^2/\sum_t\rho_t^2$ beside every reward, since weights worth a fraction of
their count have shown the block little whatever the surrogate says.

Algorithm~\ref{alg:pmpo} gives CDPO in full. The recipe runs in four steps. Episodes are collected under the policy that is acting. A price is drawn per decision, and an episode's return is its terminal reward discounted by the escalations it made at that price. The advantage is that return against the mean return over episodes at the same prices, shared by every decision of the episode. The block is then updated by the clipped surrogate of PPO on the per-decision importance ratio.

\begin{algorithm}[t]
\caption{CDPO: updating the self-routing block from episode returns}
\label{alg:pmpo}
\small
\begin{algorithmic}[1]
\REQUIRE acting policy $\mu$; block $H_\theta$, backbone frozen, giving $v_t$ by equation~\ref{eq:gain}; prices $[\lambda_{\min},\lambda_{\max}]$; temperature $\tau$; clip $\epsilon$; step $\eta$
\STATE $\triangleright$ \textit{Rollout, once: the data is off-policy from here on}
\STATE run $\mu$ over each episode, recording the state $x_t$ and escalation $e_t\in\{0,1\}$ of every decision and the terminal return $R$
\FOR{each epoch}
  \STATE $\triangleright$ \textit{Return}
  \STATE $\lambda_t\sim\mathcal{U}[\lambda_{\min},\lambda_{\max}]$, one a decision \COMMENT{redrawn, so one block serves every price}
  \STATE $G_t\leftarrow R\exp\big(-\lambda_t\textstyle\sum_{s} e_s\big)$ \COMMENT{the episode's return, discounted at this decision's own price}
  \STATE $A_t\leftarrow G_t-\bar{G}$, for $\bar{G}$ the mean of $G$ over the batch at these prices
  \FOR{each minibatch of $n$ decisions}
    \STATE $\triangleright$ \textit{Loss}
    \STATE $\pi_\theta(e_t\mid x_t,\lambda_t)\leftarrow\sigma\big((v_t-\lambda_t)/\tau\big)$ for $e_t=1$, one minus it for $e_t=0$
    \STATE $\rho_t\leftarrow\pi_\theta(e_t\mid x_t,\lambda_t)\,/\,\mu(e_t\mid x_t)$ \COMMENT{per decision; along an episode they collapse}
    \STATE $\mathcal{L}\leftarrow-\frac{1}{n}\sum_t\min\big(\rho_t A_t,\ \mathrm{clip}(\rho_t,1-\epsilon,1+\epsilon)\,A_t\big)$ \COMMENT{equation~\ref{eq:cdpo}}
    \STATE $\triangleright$ \textit{Update}
    \STATE $\theta\leftarrow\theta-\eta\nabla_\theta\mathcal{L}$ \COMMENT{AdamW; no critic}
  \ENDFOR
\ENDFOR
\ENSURE the updated block $H_\theta$
\end{algorithmic}
\end{algorithm}

\clearpage
\onecolumn
\section{Routing case studies on JevBench}
\label{app:cases}

Seven of JevBench's $231$ public items, taken from the served $\lambda=0$ run on
Qwen3.8-27B at $B=4096$ that Table~\ref{tab:jevbench} reports, with the block it
deploys. One run supplies all three readings of each item: what the fast mode
answered, what the slow mode answered and at what cost, and which of the two the
block chose. At $\lambda=0$ it escalates $124$ of the $231$, and every case below
is shown at that price. The state, the instruction and the options are given in
full, and a state carrying a table or a structured object is shown both as it
renders and as the model received it; only the slow mode's reasoning is abridged.

In the first four the fast mode is wrong, the slow mode is right, and the block
pays for the slow pass. What it pays varies by an order of magnitude across them,
from $326$ tokens to $3{,}415$.

\par\medskip{\centering\includegraphics[width=\textwidth]{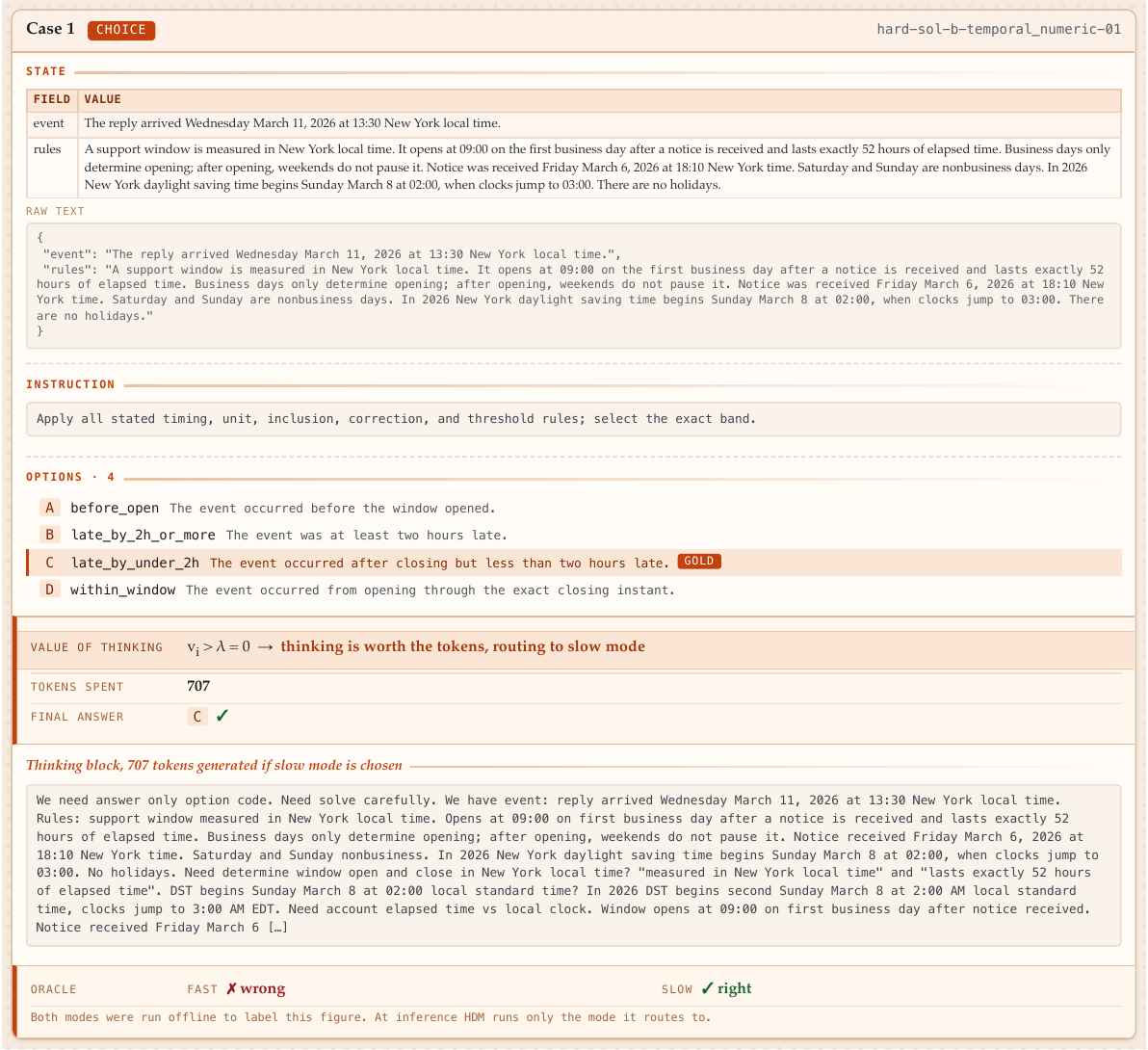}\par}\medskip
\par\medskip{\centering\includegraphics[width=\textwidth]{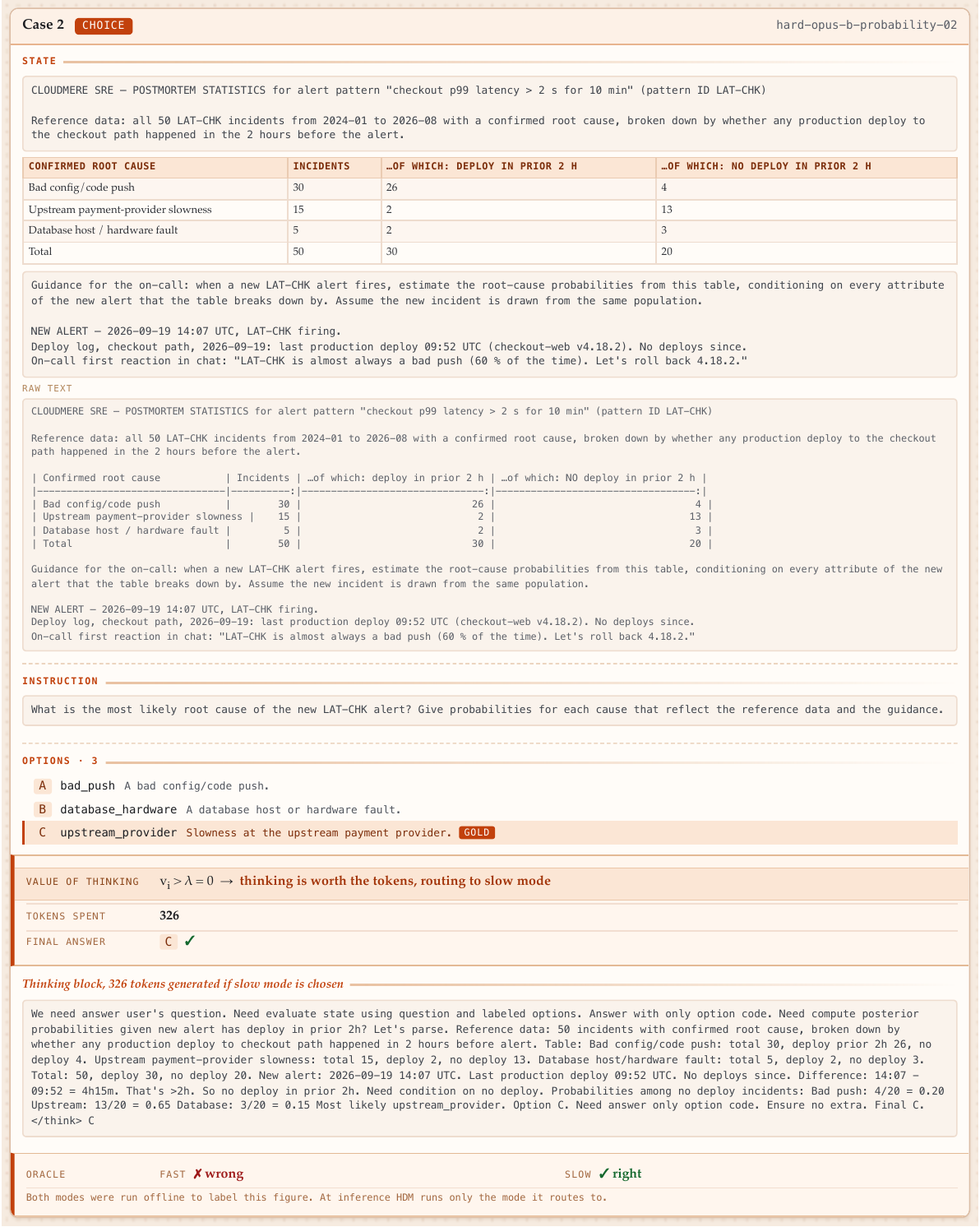}\par}\medskip
\par\medskip{\centering\includegraphics[width=\textwidth]{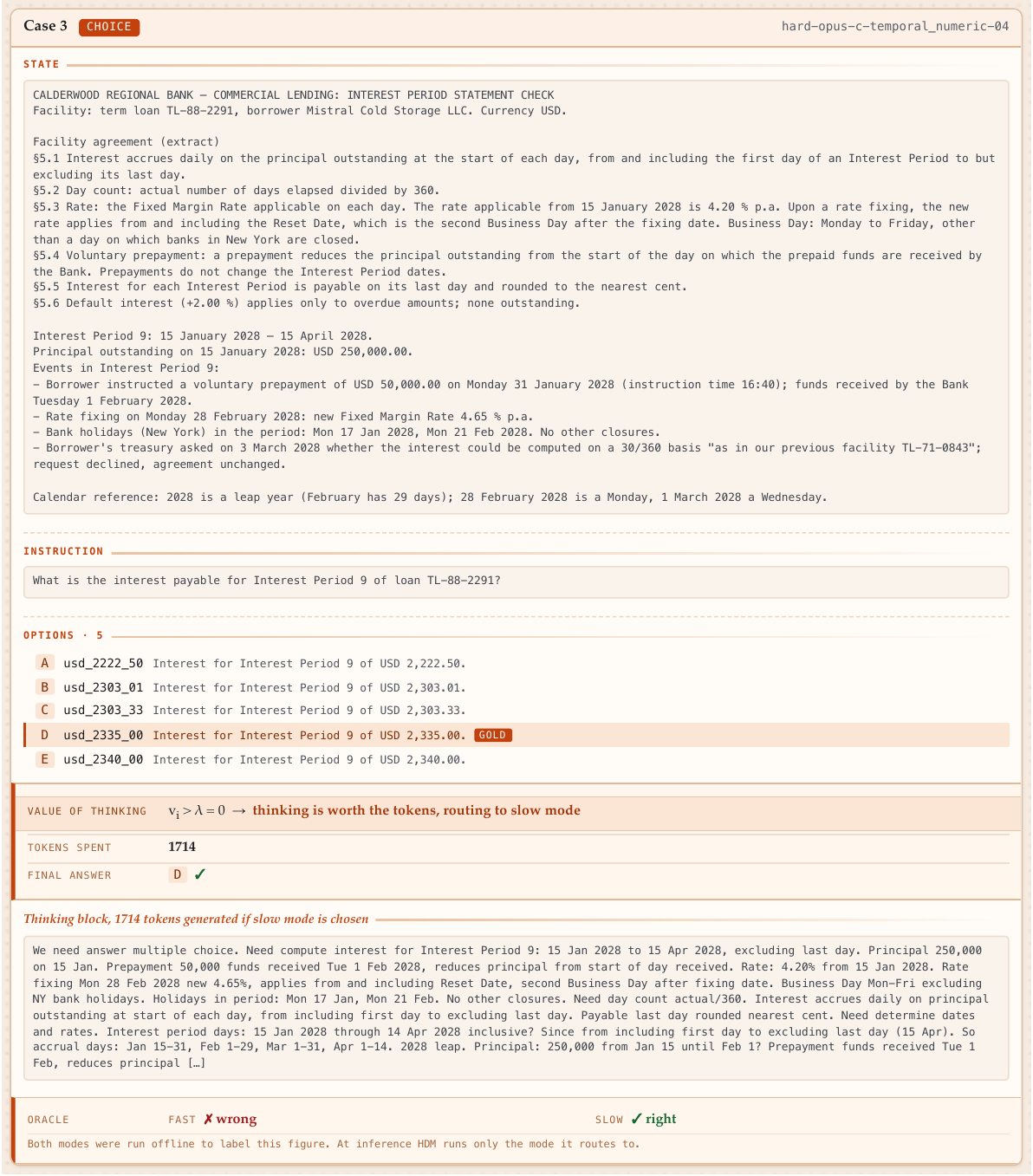}\par}\medskip
\par\medskip{\centering\includegraphics[width=\textwidth]{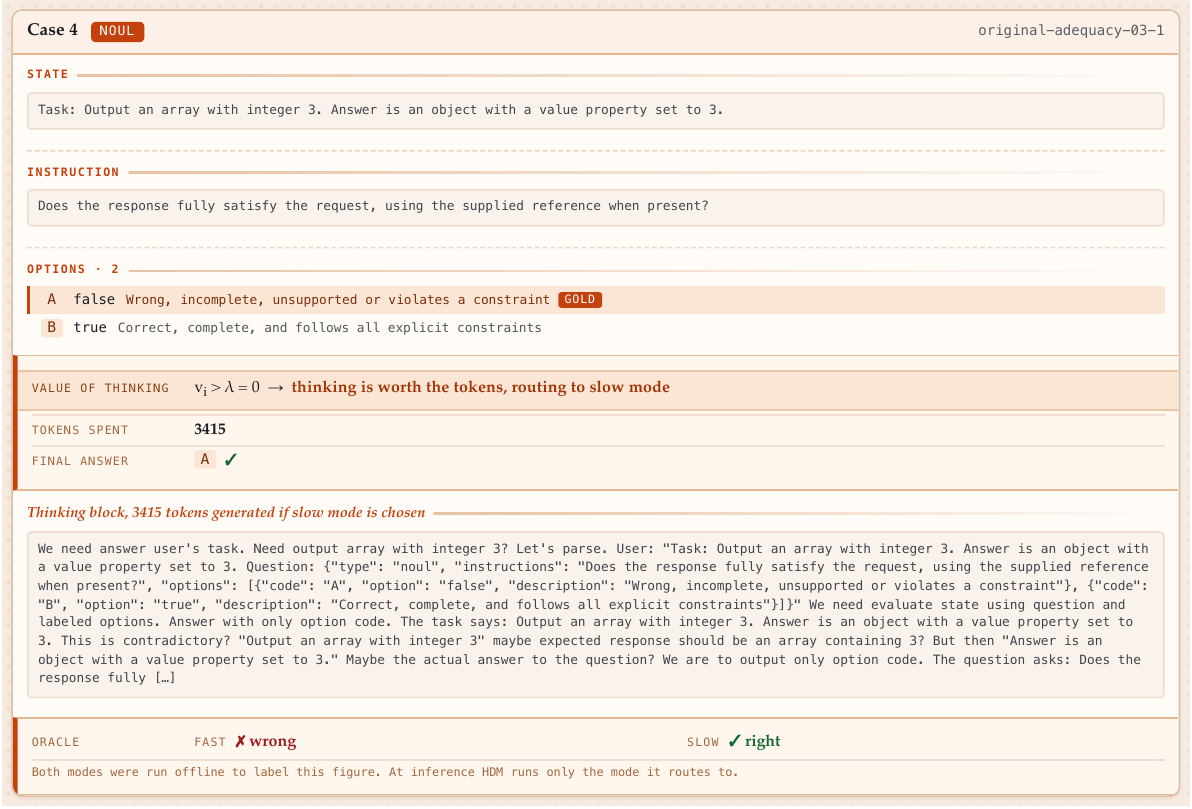}\par}\medskip

In the last three both modes answer correctly, so the thinking tokens listed are
what the block declined to spend, $1{,}523$ on the first and $224$ on the last.

\par\medskip{\centering\includegraphics[width=\textwidth]{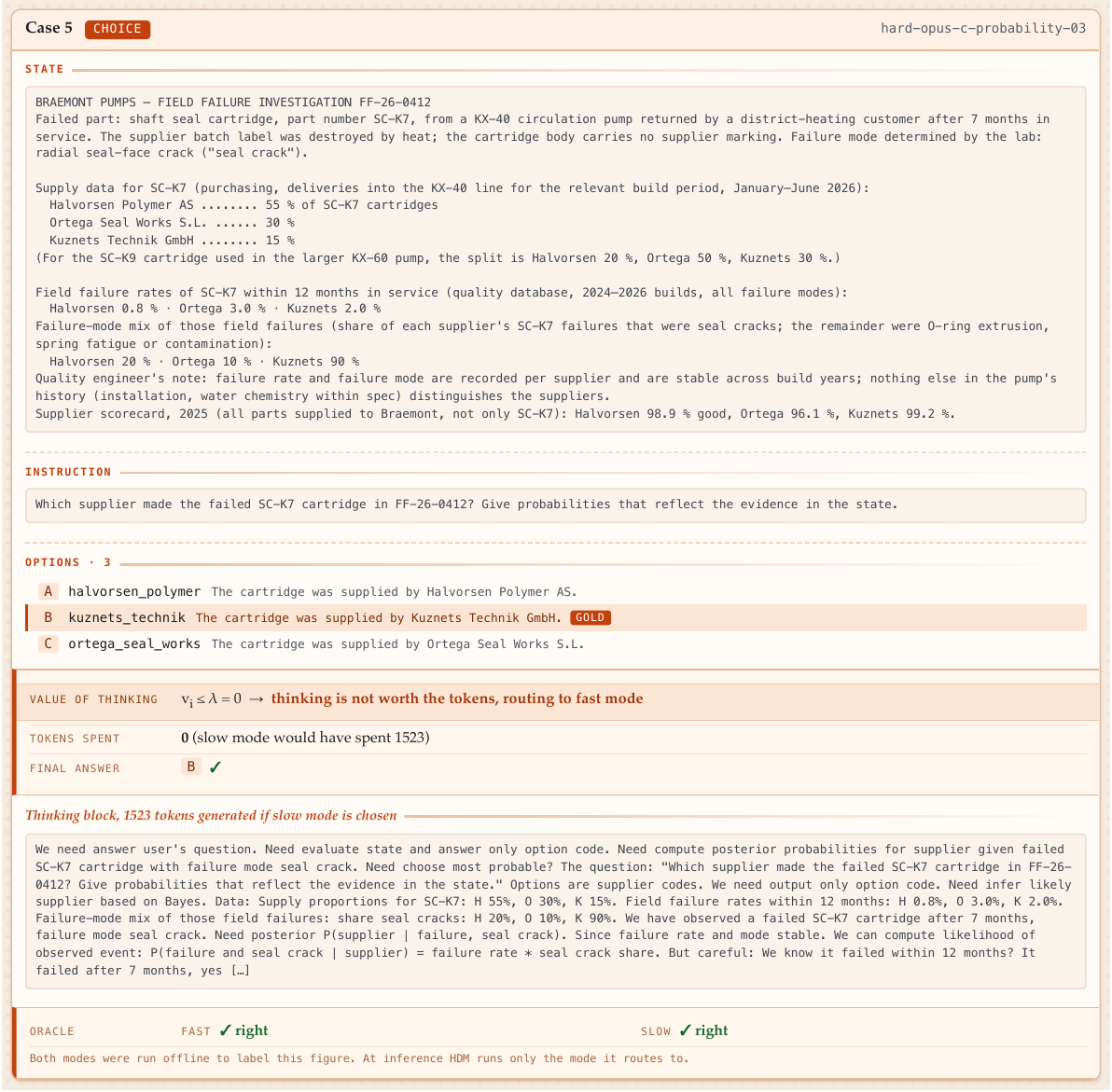}\par}\medskip
\par\medskip{\centering\includegraphics[width=\textwidth]{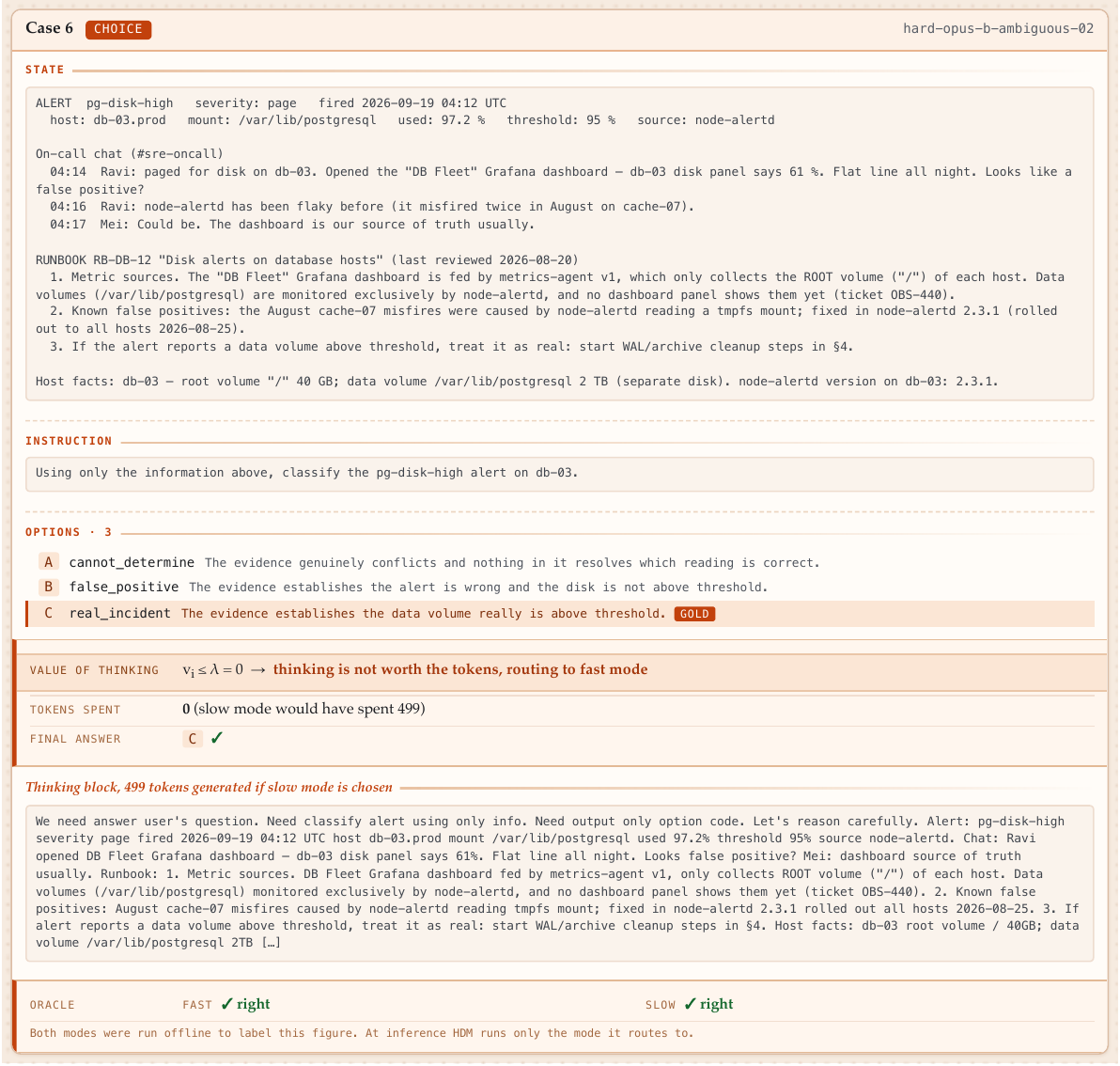}\par}\medskip
\par\medskip{\centering\includegraphics[width=\textwidth]{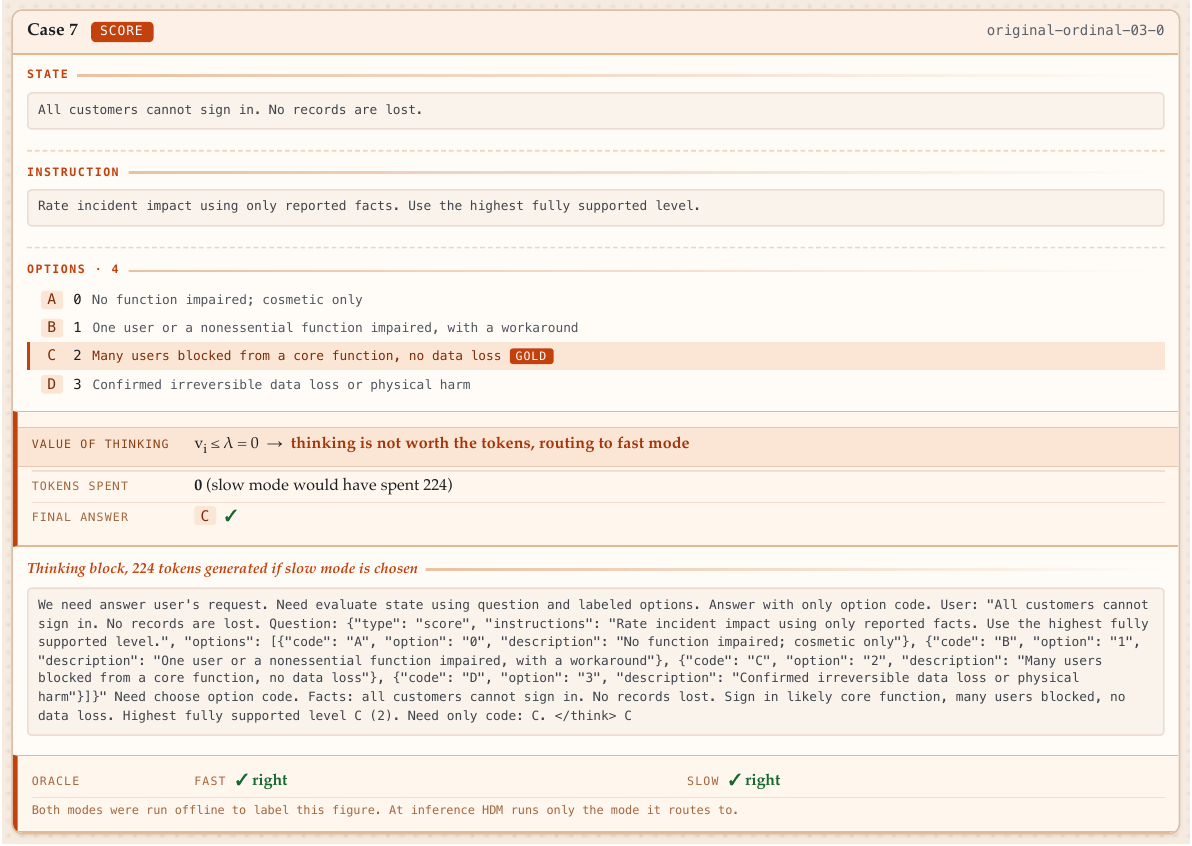}\par}\medskip

\end{document}